\documentclass{article}

\usepackage[final]{corl_2018}

\usepackage{graphicx}
\usepackage{algorithm, algorithmicx, algpseudocode}

\title{NavigationNet: A Large-scale Interactive Indoor Navigation Dataset}

\author{
  He Huang \\
  Department of Computer Science and Engineering\\
  Shanghai Jiao Tong University
  P.R. China\\
  \texttt{hjhhh3000@sjtu.edu.cn} \\
  \And
  Yujing Shen \\
  Department of Computer Science and Engineering\\
  Shanghai Jiao Tong University
  P.R. China\\
  \texttt{yujingshen96@gmail.com} \\
  \And
  Jiankai Sun \\
  Department of Computer Science and Engineering\\
  Shanghai Jiao Tong University
  P.R. China\\
  \texttt{jiankai@sjtu.edu.cn} \\
  \And
  Cewu Lu (Correponding Author)\\
  Department of Computer Science and Engineering\\
  Shanghai Jiao Tong University
  P.R. China\\
  \texttt{lucewu@sjtu.edu.cn} \\
}

\begin{document}
\maketitle

%===============================================================================

\begin{abstract}
Indoor navigation aims at performing navigation within buildings. In scenes like home and factory, most intelligent mobile devices require an functionality of routing to guide itself precisely through indoor scenes to complete various tasks in order to serve human. In most scenarios, we expected an intelligent device capable of navigating itself in unseen environment. Although several solutions have been proposed to deal with this issue, they usually require pre-installed beacons or a map pre-built with SLAM, which means that they are not capable of working in novel environments. To address this, we proposed NavigationNet, a computer vision dataset and benchmark to allow the utilization of deep reinforcement learning on scene-understanding-based indoor navigation. We also proposed and formalized several typical indoor routing problems that are suitable for deep reinforcement learning.
\end{abstract}


%===============================================================================

\section{Introduction}
\label{sec:introduction}
Indoor navigation aims at preforming navigation within buildings, such as train stations, airports, shopping centers, offices and museums. Most intelligent devices require an indoor routing functionality to guide itself precisely through rooms to complete various tasks. In most of the scenarios, we expect an intelligent device capable of navigating itself in new environments to serve human.

In the past years, several solutions have been proposed to deal with this problem \cite{HLiu2007IEEESMC}. Although they can be applied in some special cases, none of them can work smartly in novel environments or are based on scene understanding. Those solutions require either dense pre-installed beacons (e.g. WiFi/Bluetooth chips) over the scene or a map pre-built with SLAM in which human workers are required to carry cameras to scan the scene in advance. Such requirements significantly limit the applications of these solutions. New scenes have to be well prepared before the robots can work properly in it, which makes it impossible to be put into large-scale application in various indoor scenarios. Additionally, neither beacons nor maps are helpful in semantically understanding indoor scenes and thus will lead to the failure of the robot to perform many practical tasks. For example, to complete a regular command like 'to take a cup', a housekeeping android should understand what the scene is like, what a cup is and where a cup usually be, to navigate properly.

We expected a more general solution of indoor navigation problems that requires no preparation of the scene in advance. Such a high demand requires a more human-like robot that is much more intelligent to implement self-exploration in novel environments. Recent years, we have witnessed the rapid development of deep reinforcement learning (DRL) \cite{VMnih2013CoRR,VMnih2016CoRR} and computer vision \cite{KAlex2012NIPS,KHe2015CoRR}. Therefore, smart indoor navigation enabled by DRL with visual inputs has been introduced \cite{YZhu2016CoRR}. In this setting, pre-installed beacons or pre-built map are no longer required. A smart robot with vision will be enough to do the navigation job in different scenes.

To apply reinforcement learning methods on navigation problems, massive trial-and-error is necessary in training the model. A straight-forward way is to construct a real robot and train Reinforcement Learning (RL) models on real-world scenes. However, such process is extremely slow and costly. Additionally, it is often accompanied by robot collision and damage, which means that more time and resources would be wasted. Even worse, due to the bottleneck of physical movements, to execute any action in the real world requires at least several seconds, which is unacceptable for any machine learning methods since that they usually require millions of times of trial-and-error. Aside from those drawbacks, such a process raises the problem of reproducibility. Even if all the background settings of a research are published in detail, it is still very hard for third parties to reproduce the experiment result since it is costly and time-consuming to find or construct a new environment that is identical to the one the original author used. That makes it impossible for third parties to examine and evaluate a new research result.

We believe, an open-source, low-cost, large-scale dataset and corresponding benchmarks would be a key to largely advance this research field. Some previous work has tried to train the robot in computer graphics (CG) scenes like AI2-THOR\cite{YZhu2016CoRR}, SUNCG\cite{SSong2016CVPR}. However, we do have noticed that training models on CG environments raised the problem of model transferring as the gap between hand-crafted scenes and the real world is significant. The visual appearance of images rendered by CG system always look non-realistic. Existing 3D models are limited in representing real-world scenes and objects.

To deal with this problem, we proposed NavigationNet, a large-scale, real-world, interactive dataset. In each scene, we use human-size robot with 8 cameras to capture photos towards 8 different directions at all walk-able positions. Given all the collected images, we can easily tour the room like a CG system by moving among those walk-able positions. However, different from CG systems, what we observed from this dataset are all real images captured from the real world. Therefore, we can virtually command a robot to walk in the scene and get the first-perspective view of the robot. Our dataset is interactive. That is to say, as we send a command of movement (e.g. moving forward, turning right), the robot will perform as required and return a new view it sees from the new position. To our best knowledge, our NavigationNet is the first real-world indoor scene dataset that is interactive in computer vision field. Our dataset covers about 1500 $m^2$ indoor area with various bedrooms, studies, meeting room and etc. This makes the bias less in experiments on our dataset.

Based on NavigationNet, we proposed four applications that require indoor navigation. We will explained them in detail in chapter \ref{sec:applications}.

To summarize, we presented a large-scale dataset consisting of visual data collected from real scenarios to allow a low-cost, reproducible training of indoor navigation robots based on reinforcement learning methods. We also proposed and formalized several practical indoor navigation tasks in the framework of reinforcement learning.

%===============================================================================

\section{Related Work}
\label{sec:related_work}
In this chapter, we will discuss some of the previous works that are related to our work or act as a tool in this project. First we will talk about the technology of Visual Semantic Planning. A brief introduction to the popular datasets in the field of computer vision, after which we will explain three techniques used in our work in detail: Reinforcement Learning, Convolutional Neural Network(CNN) and Simultaneous Localization and Mapping.

\subsection{Visual Semantic Planning} 
\label{subsec:related_work/visual_semantic_planning}
Fundamental tasks in mobile robot controlling include navigation, mapping, grasping, path planning and so on. Such task of interacting with a visual world and planning a sequence of actions to achieve a certain goal is addressed as \emph{Visual Semantic Planning} \cite{YZhu2017ICCV}. A series of work \cite{EJang2017CoRR, SGupta2017CoRR, SHuang2016IEEETIP, CFinn2015CoRR} has addressed the task-level planning and learning from dynamics.  Classical approaches \cite{JCanny1988CRMP, SLaValle2000RERT} of motion planning require building map purely geometrically using Light Detection and Ranging (LiDAR) or Structure from Motion (SfM) \cite{JSchonberger2016CVPR, JSchoenberger2016ECCV, MPollefeys2004IJCV}. Srivastava \emph{et al.} \cite{SSrivastava2014ICRA} provides an interface between task and motion planning, such that the task planner can effectively operate in an abstracted state space that ignores geometry. Recently proposed end-to-end learning-based approaches \cite{YZhu2017ICRA,SGupta2017CoRR,SLevine2016JMLR} can go directly from pixels to actions. For instance Zhu \emph{et al.} \cite{YZhu2017ICRA} adopt feed-forward neural network to target-driven visual navigation in indoor scenes. Gupta \emph{et al.} \cite{SGupta2017CoRR} use online Cognitive Mapping and Planning (CMP) approach for goal direction visual navigation without requiring a pre-constructed map. Jang \emph{et al.} \cite{EJang2017CoRR} train a deep neural network to perform the semantic grasping task inspired by the "two-stream hypothesis" of human vision. In terms of Complete Path Planning, Kollar \emph{et al.} \cite{TKollar2008IJRR} use reinforcement learning to find trajectories and the learnt policy transfers successfully to a real environment.

\subsection{Datasets for Computer Vision} \label{subsec:related_work/datasets_for_computer_vision}
Datasets and corresponding benchmarks have played a significant role in many areas such as computer vision and speech recognition. The evolution from WordNet\cite{CFellbaum1998WN} to ImageNet\cite{ORussakovsky2014CoRR} was one of the many successful examples that proved the power of an effective dataset to accelerate the development of one area. On the contrary, the lack of appropriate data has become one of the most crucial challenges for deep reinforcement learning, especially when it is dealing with real world problems like visual semantic learning. Since training and quantitatively evaluating DRL algorithms in real environments is either impractical or costly, this problem has become more urgent. For RL algorithms dealing with virtual-world problems, the Arcade Learning Environment (ALE) \cite{MBellemare2012CoRR} exposed Atari 2600 games as reinforcement learning problems. Works such as Duan \emph{et al.} \cite{YDuan2016CoRR} and Brockman \emph{et al.} \cite{GBrockman2016arXiv} propose toolkits to qualify progress in reinforcement learning. Other benchmarks \cite{BTanner2009JMLR, AGeramifard2015JMLR,GSynnaeve2016arXiv,CBeattie2016arXiv,MKempka2016CIG}, simulators \cite{BWymann2014TORCS} or physics engines \cite{ETodorov2012IROS} are also designed for the development of deep reinforcement learning. Also, scientists are trying to model the real world for DRL or other visual tasks. Chang \emph{et al.} \cite{AChang20173DV} provides 90 building-scale reconstructed scenes for supervised and self-supervised computer vision tasks such as keypoint matching, view overlap prediction semantic segmentation, scene classification and so on. AI2-THOR\cite{YZhu2017ICRA}, SUNCG\cite{SSong2016CVPR} and House3D are all CG scenes designed especially for deep reinforcement learning. The advanced version of AI2-THOR\cite{EKolve2017arXiv} even provides the functionality to let the robot directly interact with the objects in the scenarios. Different from our NavigationNet, these datasets are either synthetic or reconstructed using computer graphics techniques, which introduce inconsistency distribution between the real-world scenarios and the produced ones.

\subsection{Reinforcement Learning} \label{subsec:related_work/reinforcement_learning}
Reinforcement learning (RL) \cite{RSutton1998RL} method was proposed in the late 90s. It provides a technique to allow the robot or any other intelligent systems to build a value function to evaluate policies in given situations from an interactive way of trail-and-error. Pioneering works from Mnih \emph{et al.} \cite{VMnih2015Nature}, Lillicrap \emph{et al.} \cite{TLillicrap2015CoRR}, Schulman \emph{et al.} \cite{JSchulman2015CoRR} and Silver \emph{et al} \cite{DSilver2016Nature} introduced and ignited the idea of join Reinforcement Learning and Deep Learning into Deep Reinforcement Learning.

In learning complex behavior skills and solving challenging control tasks in high-dimensional raw sensory state-space, DRL methods have shown tremendous success \cite{DSilver2016Nature,JSchulman2015CoRR2,VMnih2016CoRR,YWu2017CoRR}. Trust Region Policy Optimization (TRPO)  \cite{JSchulman2015CoRR2} makes a series of approximations to the theoretically-justified procedure and has robust performance on a wide range of tasks.
Asynchronous Advantage Actor-Critic (A3C) Algorithm \cite{VMnih2016CoRR}, which uses asynchronous gradient descent for optimization, succeeds on task of navigating random 3D mazes as well as a wide variety of continuous motor control problems.

UNsupervised REinforcement and Auxiliary Learning (UNREAL) \cite{MJaderberg2016CoRR} brings together the A3C framework with auxiliary control tasks and auxiliary reward tasks. Actor Critic using Kronecker-factored Trust Region (ACKTR) \cite{YWu2017CoRR} Algorithm uses a Kronecker-factored approximation to natural policy gradient that allows the covariance of the gradient to be inverted efficiently. Proximal Policy Optimization (PPO) \cite{JSchulman2017CoRR} Algorithms use multiple epochs of stochastic gradient ascent to perform each policy update. Reinforcement Learning (RL) provides a powerful and flexible framework to several applications. Andrew Y. Ng \emph{et al.} \cite{HKim2004NIPS} describe a successful application of autonomous helicopter.
Finn \emph{et al.} \cite{CFinn2015CoRR} present an approach that automates state-space construction by learning a state representation directly from camera images with deep spatial autoencoder. Heess \emph{et al.} \cite{NHeess2017CoRR} explore a rich environment can help to promote the learning of complex behavior. They normalize observations, scale the reward and use per-batch normalization of the advantages. Gu \emph{et al.} \cite{SGu2017ICRA} demonstrate a DRL algorithm based on off-policy training of deep Q-function can scale to complex 3D manipulation tasks and can learn deep neural network policies efficiently enough to train on real physical robots. Denil \emph{et al.} \cite{MDenil2016arXiv} find DRL methods can learn to perform the experiments necessary to discover properties such as mass and cohesion of objects.

\subsection{Convolutional Neural Network} 
\label{subsec:related_work/convolutional_neural_network}
Deep convolutional neural networks \cite{AKrizhevsky2012NIPS,YLeCun1989NC} have led to a series of breakthrough for visual tasks. He \emph{et al.} \cite{KHe2015CoRR} present a Residual Network (ResNet) to ease the training of deep neural network. We extract feature map with ResNet 101. Huang \emph{et al.} \cite{GHuang2016CoRR} presents Dense Convolutional Network (DenseNet) which connects each layer to every other layer in a feed-forward fashion. Chen \emph{et al.} \cite{YChen2017CoRR} propose Dual Path Net (DPN) by revealing ResNet and DenseNet within the higher order recurrent neural network framework. Deep Learning models have been successful in learning powerful representations and understanding stereo context \cite{JPang2017arXiv,JZbontar2016JMLR}. Shaked \emph{et al.} \cite{AShaked2016arXiv} present a three-step pipeline for the stereo matching problem and a network architecture (ResMatch) for computing the matching cost at each possible disparity. Kendall \emph{et al.} \cite{AKendall2017arXiv} propose an end-to-end model (GC-Net) for regressing disparity from a rectified pair of stereo images.

\subsection{Simultaneous Localization and Mapping} 
\label{subsec:related_work/simultaneous_localization_and_mapping}
Visual SLAM plays an important role in autonomous navigation of mobile robots \cite{JFuentesPacheco2015AIR}. The emergence of MonoSLAM \cite{ADavison2003ICCV} makes the idea of utilizing one camera become popular. Parallel Tracking and Mapping (PTaM) \cite{GKlein2007MAR} uses an approach based on keyframes with two parallel processing threads. ORB \cite{ERublee2011ICCV} becomes a computationally-efficient replacement to SIFT keypoint detector and descriptor that has similar matching performance. Keyframe bundle adjustment outperforms filtering, since it gives the most accuracy per unit of computing time \cite{HStrasdat2012IVC}. S-PTAM \cite{TPire2015IROS} allows accuracy improvement of the mapping process with respect to monocular SLAM and avoiding the well-known bootstrapping problem. ORB-SLAM2 \cite{RMurArtal2016arXiv} is a complete SLAM system for monocular, stereo and RGB-D cameras. RGB-D SLAM \cite{TWhelan2015IJRR} can producing high quality globally consistent surface reconstruction with only a low-cost commodity RGB-D sensor. Mur-Artal \emph{et al.} \cite{RMurArtal2017RAL} propose a visual-inertial monocular SLAM which is able to close loops and reuse its map to achieve zero-drift localization in already mapped areas.

%===============================================================================

\section{NavigationNet}
\label{section:navigationnet}
NavigationNet is specifically designed for applying reinforcement learning methods to indoor navigation tasks. We collected data from the real world and organized it properly to allow the robot to roam in the dataset as if they are in the corresponding real-world scenario. In this chapter, we will explain its organization and applications in detail to show you the world of NavigationNet.

\subsection{Data Organization}
\label{subsec:navigationnet/data_organization}
\begin{figure*}
\begin{center}
\fbox{\includegraphics[width=\textwidth]{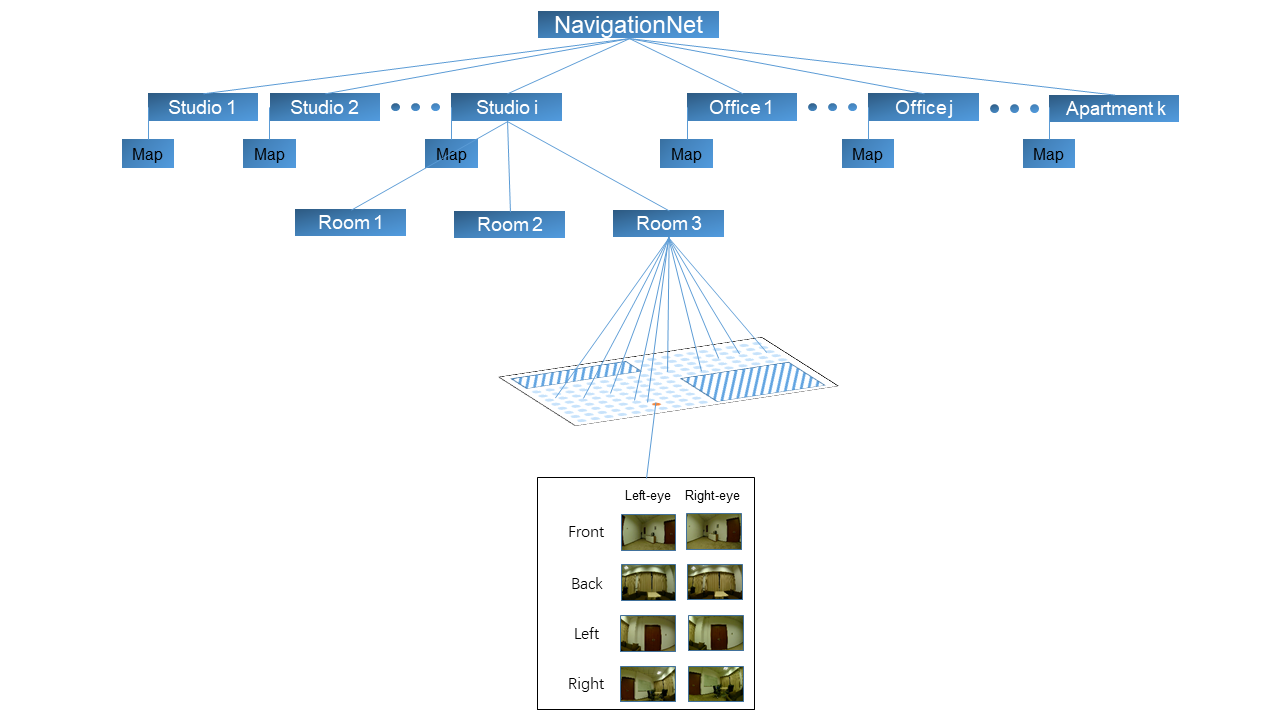}}
\end{center}
   \caption{NavigationNet Structure}
\label{fig:navigationnet/data_organization/navigationnet_structure}
\end{figure*}

Images in NavigationNet are organized hierarchically. The hightest node is the \emph{root} of NavigationNet. The second level nodes, also the primary elements, are \emph{scenes}. A \emph{scene} is a collection of images collected from the same indoor space. In the next level, a \emph{scene} is divided into \emph{rooms}, space connected with doors. In the current version of NavigationNet, we have 15 scenes, each with 1-3 rooms. The origin room for each scene is at least 50 $m^2$ in area.

The fourth level is called \emph{postion}. A \emph{position} is from where images are collected. A \emph{room} contains hundreds of \emph{postions}. When being trained, the robot could move among the adjacent \emph{positions} to 'see' from that perspective. When constructing the dataset, our mobile robot iterated over all the walkable points in the room with the granularity of 20cm. Hence, in the dataset, two adjacent \emph{positions} are 20cm away from each other. With such a granularity, one room usually contains thounsands of \emph{positions}.

From the perspective of information, a \emph{position} can be seen as a bundle of information that the robot will receive when it is placed at that \emph{position}. Usually, we offer 8 images in this bundle. These images are divided into four directions: front, back, left and right. Towards each direction, two parallel images are taken to complete a binocular vision system. Such a setting enables the possibilities like stereo matching and panorama reconstruction.

In addition to the images, we also provide a ground-truth map attached to the scene. The map can be considered as a binary map indicating where is walkable or non-walkable. This is essential especially to the Auto-SLAM problem.

\subsection{Robot Moving Control}\label{subsec:navigationnet/robot_moving_control}
To satisfy the different requirements of potential tasks, we tried our best to provide a more generalizable moving control SDK for NavigationNet.

As is stated in section \ref{subsec:navigationnet/data_organization}, the fundamental element in NavigationNet is \emph{scene}, virtual robots are allowed to roam in scenes. To serve well as an action in a reinforcement learning task, movements should be standard, discrete, limited and complete.

\begin{itemize}
    \item An action should be standard, so that the robot should move the same distance towards the same direction or turning the same angle when a specific instruction of a movement is given.
    \item An action should be discrete, so that the action space can be finite and discrete.
    \item Actions should be limited, so that the algorithm do not need to choose actions from an infinite action space.
    \item Actions should be complete, so that all the actions joined together could describe the whole real-world action space.
\end{itemize}

In practice, it is impossible to use a finite action space to cover an infinite one. We need some sort of compromise. The one we take in this case is to discrete the movement length and the turning angle. So that we defined six types of movements.

\begin{itemize}
    \item \textbf{MOVE FORWARD} Ask the robot to move forward one position. For example, go from (9, 12) to (9, 13) when facing north but go from (11, 8) to (10, 8) when facing west. Since two adjacent nodes are 20cm away, this is to move forward 20cm in real-world scenes.
    \item \textbf{MOVE BACKWARD} Ask the robot to move backward one position. This is the opposite movement of MOVE FORWARD and can revert the effect of MOVE FORWARD.
    \item \textbf{MOVE LEFT} Ask the robot to move left one position. For example, go from (5, 4) to (4, 4) when facing north but go from (7, 4) to (7, 3) when facing west. The real world distance would be the same as MOVE FORWARD and MOVE BACKWARD. It should be paid attention to that MOVE LEFT is very different from TURNING LEFT as it will move the position of the robot but will not change the facing direction.
    \item \textbf{MOVE RIGHT} Ask the robot to move right one position. This is the opposite movement of MOVE LEFT and can revert the effect of MOVE LEFT.
    \item \textbf{TURN LEFT} Ask the robot to turn left at 90-degree. For example, when the robot receives this request when at (5, 7) facing north, it should then facing west still at (5, 7). It should be paid extra attention that this movement is different from MOVE LEFT that the former changes the facing while the latter change the position.
    \item \textbf{TURN RIGHT} Ask the robot to turn right at 90-degree. This is the opposite movement of TURN LEFT.
\end{itemize}

\begin{figure*}
\begin{center}
\fbox{\includegraphics[width=\textwidth]{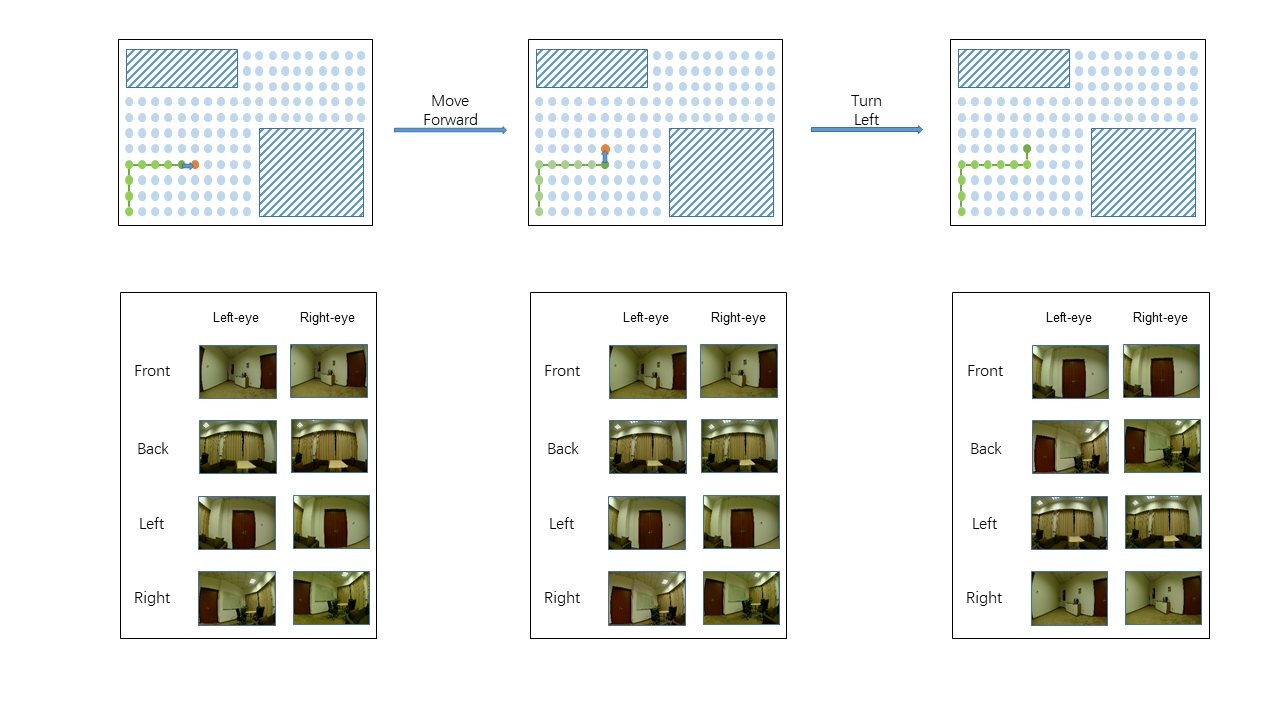}}
\end{center}
   \caption{Movements}
\label{fig:navigationnet/data_organization/movements}
\end{figure*}

At the same time, we should make it clear that, in any given tasks, according to the specific setting, not all the actions (movement) need to be taken into consideration. For example, when we would like to simulate a two-wheel one-eye robot, TURN LEFT and TURN RIGHT are a must, otherwise seventy-five percent of the images can never be perceived while MOVE LEFT and MOVE RIGHT should be eliminated since a two-wheel robot can not make the exact movement of \emph{moving left 20cm}. Meanwhile, when simulating a robot with panorama vision, we could only take MOVE LEFT and MOVE RIGHT but not TURN LEFT and TURN RIGHT as they are not necessary but can only enlarge the action space unnecessarily and make the task more difficult.

%===============================================================================

\section{NavigationNet Construction}
\label{sec:navigationnet_construction}
 NavigationNet is a large-scale dataset with hundreds of thousands of images. The large amount makes the data collection a task of impossible. What is worst is that, as to each image, it is required that the position, angle and height where it is taken should be exactly where it is expected so long that the simulation of reality will not offset. These two factors together make the task of collection much more difficult than expected. In this chapter, we will talk about how we constructed this dataset efficiently and accurately.

\subsection{Collector Mobile Robot}
\label{subsec:navigationnet_construction/collector_mobile_robot}
Robots are designed to liberate human beings from the repetitive boring work. NavigationNet is built for intelligent robots, also built with the help of intelligent robots. Collecting data for NavigationNet requires heavy labor and high accuracy but little flexibility (most of the possible conditions are under control), which is surprisingly suitable for robots.

In order to reduce the labor requirement, we exploited the possibilities of smart hardware. Our team developed a dedicated data-collecting mobile robot with \emph{Arduino Mega2560} and \emph{Raspberry Pi 3 Model B} codenamed \emph{GoodCar}. 

Arduino\cite{MBanzi2008Arduino} is an open-source physical computing platform based on a simple I/O board and a development environment that implements the Processing/Wiring language. Arduino boards can be programmed in the same name program language Arduino, which is a C-like area-specific language to receive and send signals from tens of low-voltage I/O ports. It is specifically suitable for robot controlling. In this project, we use a \emph{Arduino Mega2560} to control the movement of the robot.

Raspberry Pi\cite{GHalfacree2012RaspberryPi} is a cheap single-board computer system running a specificialy made Linux distribution. It is originally built for computer programming education but we find it suitable for robot controlling. It consumes little electricity which allows us to strip the 220V power wires. It contains all kinds of I/O ports such that we could plug it with Arduinos, cameras, other full-sized computers and many other modules to make it the center of the system. It runs standard linux distribution such that the possibility of software extension is beyond imagine. Last but not least, it is cheap so that we could us as many as we want on the robot. In practice, we use two \emph{Raspberry Pi 3B} for controlling all other modules and communicating with the upper computer.

To build this robot, we used a two-level structure. The upper-level is as high as 1.4m to simulate the the height of the eyes of an average adult. We plugged-in eight cameras on this level, two for each direction to make a stereo vision system and to avoid unnecessary turning-around.

The lower-level is for the motion systems. The Raspberry Pi and Arduino mentioned above are all on this level. The Arduino is to control the mobile devices and sensors. It is linked with four motors to control the movement. In the meantime, it is linked with a sonar and code plate counters to avoid collision and measure the distance of movement. Also it is linked with a Raspberry Pi and under its control. We used two \emph{Raspberry Pi 3B} in each robot, one is master and the other is slave. The master computer controls many things. First it communicates with the Arduino to control the movement indirectly. Second it is responsible for evaluating the signal Arduino sent back to avoid collision. Third it is master computer's duty to control four of the cameras to take photo at given position and store the images on the SD card. Fourth the slave computer is also under its control to take photos when required. Lastly it communicates with the upper computer, which is controlled by human, via Wi-Fi connection. That is the control core of our robot.

Under the lower-level is the mobile devices. We have tried on many types of mobile devices and at last we used four track structures to reduce error.

We had constructed six mobile robots before moving on to collect data.

\begin{figure*}
\begin{center}
\fbox{\includegraphics[scale=0.8]{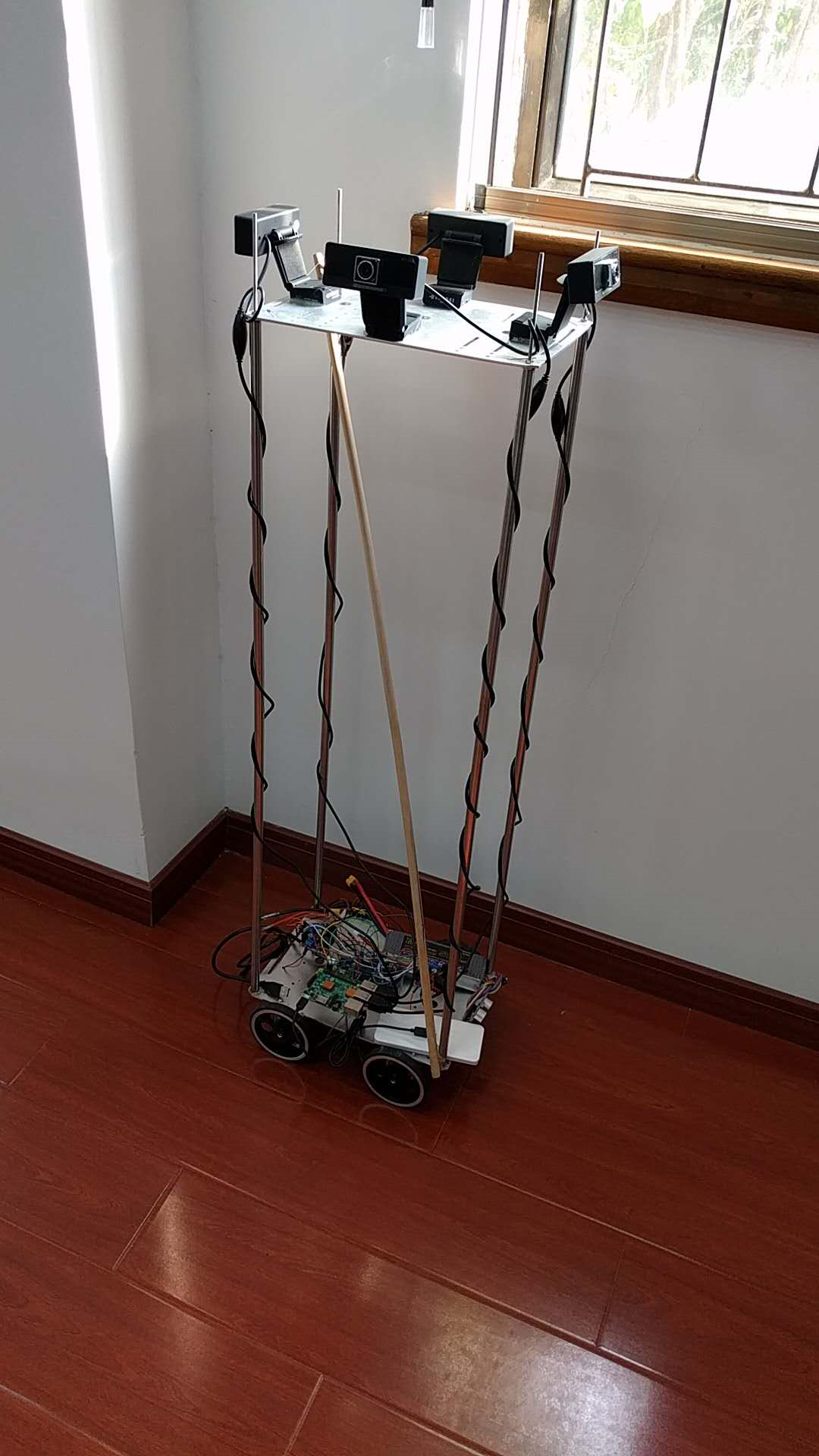}}
\end{center}
   \caption{Robot}
\label{fig:navigationnet/collector_mobile_robot/robot}
\end{figure*}

\subsection{Data Collection}
\label{subsec:navigationnet_construction/data_collection}
To make the data collecting process efficient and accurate, also to make the whole process manageable and under control, we designed and followed the following instructions:

\renewcommand{\theenumi}{\textbf{Step \arabic{enumi}}}
\begin{enumerate}
    \item Measure the size of the target room(s). Find the most southwestern corner as the starting point.
    \item Move the robot to the starting point. Evaluate and record its distance towards the borders.
    \item Let the robot face north and start to collect required photos at this point.
    \item After photos are taken, let the robot run 20cm towards north and collect data.
    \item Keep iterating over \textbf{Step 3-4} until the robot reaches the end of the line.
    \item Move the robot to the starting point of the line and then move 20cm towards east.
    \item Iterating over \textbf{Step3-6} until the robot reaches the east corner.
    \item Complete this scenario by taking photos from those points that the robot did not reach.
\end{enumerate}
\renewcommand{\theenumi}{\arabic{enumi}}

\subsection{Data Processing}
\label{subsec:navigationnet_construction/data_processing}
After the collection, the original data should be pre-processed before taken into production. The data processing steps we taken includes:

\begin{itemize}
    \item \textbf{Data Cleaning} When collecting data with robot, there are always inevitable errors occurring especially when dealing with room corners. For each room we collect, we will scan over all the collected photos and find errors. If necessary, we would go and retry collecting data to complete the set.
    \item \textbf{Map Building} When collecting data, we also recorded whether a point is accessible. With this information, we are able to rebuild a walkable-or-not map for reference.
    \item \textbf{Stereo Vision} For some of the task, to simulate a two-eye system or a RGB-D system, it would be of much help if we could produce depth channel in the very beginning. To serve this purpose, we use MC-CNN\cite{JZbontar2016JMLR} to produce some pre-processed depth channel data.
\end{itemize}

%===============================================================================

\section{Applications}
\label{sec:applications}
NavigationNet is designed for indoor navigation problems. In this chapter, we would introduce four applications base on our dataset and formalize them with the idea of reinforcement learning. Apart from these four applications, we believe our dataset will spawn more applications.

\subsection{Target-driven Navigation}
\label{subsec:applications/targetdriven_navigation}

\paragraph*{Motivation} In the real scenarios, as for robots, the key problem of indoor navigation is not about going to some place but finding some specific item. For example, as a home android, we do not ask it to get to the kitchen safely but to fetch the milk for the master. Also, as a robot in a warehouse, the ability to go to some specific location is never more important than finding a given good. To achieve this goal, a poorly trained robot control system may have to traverse over everywhere it can reach to search target object, which is neither safe nor efficient, since it has no prior knowledge like milk should be in the kitchen, or tables are usually aligned to the wall. This usually falls in the field of visual semantic learning. That is to say, to be well prepared for this kind of tasks, the robots should be able to understand the scene, knowing, for example, what the object it perceived is.

\begin{figure*}
\begin{center}
\fbox{\includegraphics[scale=1.0]{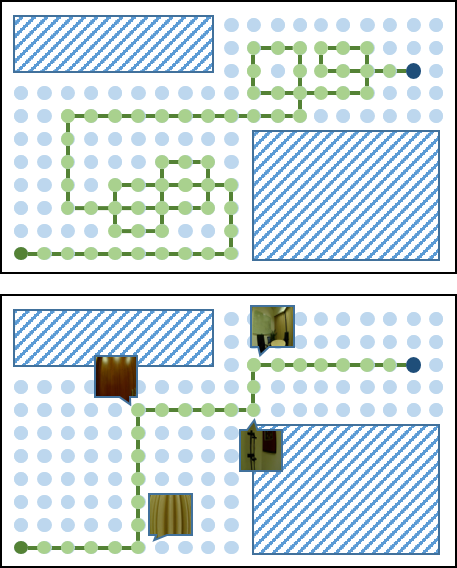}}
\end{center}
   \caption{Target-driven Navigation(Upper: Random walking, Lower: DRL-based)}
\label{fig:applications/targetdriven_navigation/targetdriven_navigation}
\end{figure*}

\paragraph*{Formalization} The formalization of this issue is inspired by Yuke Zhu \emph{et al.}\cite{YZhu2016CoRR}. To complete the task of target-driven navigation, a robot control system is given two images. One is the target image, the other is the current first-person perspective image perceived from the environment. The robot can choose to walk forward, backward, turn left or right according to the input images. The environment will update the states accordingly and give out a reward. This process loops until the target image is the same with the perceived one.

To formalize that:
\begin{itemize}
    \item \textbf{Goal} The goal is straight-forward, to find the given object. In practice, it can be converted into make the perceived image and the given image identical.
    \item \textbf{Reward} Different from the the goal, the design of the goal is not that straight-forward and requires more discussion. One thing that is sure is that a large positive reward will be given when the goal is achieved. To improve time efficiency, each step it taken should be given a small negative reward. In this paper, we use the setting of one large positive reward at the goal and minor negative in each step.
    \item \textbf{Action Space} To find an object, the understanding of the scene is the most important factor of the task. Thus we would like to simulate a one-eye robot in this problem and the Action space would be defined as \textbf{\{ MOVE FORWARD, MOVE BACKWARD, TURN LEFT, TURN RIGHT \}}
    \item \textbf{State Space} In this problem, we need not only the current perception as input but also the target image. Hence the State space would be the set of all possible image pairs.
\end{itemize}

\paragraph*{Evaluation} To evaluate such a system, many aspects should be taken into consideration. The most important two are, trajectory should be short and the robot should not run into the obstacles. Collisions should be limited strictly as only one collision could greatly damage the robot. To make a trajectory finite, we consider a trajectory longer than 10000 a loop. To calculate a overall score on all the test results, we utilize the idea of histogram. A success trajectory with $n$ steps and $m$ is a vote weighting $max(1-m/10, 0)$ to $n$. A overall score is calculated by finding the average over this histogram.

\subsection{Multi-target Navigation}
\label{subsec:applications/multitarget_navigation}
\paragraph*{Motivation} The target-driven navigation task is focused on finding one object efficiently and safely. However, in practical scenarios, this is often not enough. In many cases, we need to find more than one items and we do not care about the order. For example, in the last example about collecting goods from the different rooms and then putting them all into a box, although it is required that the robot should get to the box in the very end, we do not really care about which item it goes to get first. This has become a rather different problem since it usually requires the knowledge of adjacency. For example, milk is usually placed near breads while books are usually far away, so although the item order in the list maybe  milk, books and breads, robot should take reads instead of books first to advance the efficiency.

\begin{figure*}
\begin{center}
\fbox{\includegraphics[width=\textwidth]{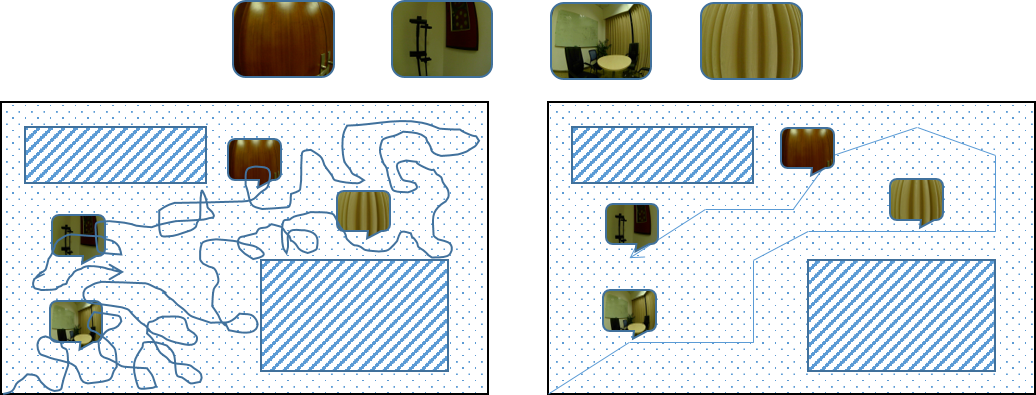}}
\end{center}
   \caption{Multi-target Navigation (Left: Random walking, Right: DRL-based)}
\label{fig:applications/multitarget_navigation/multitarget_navigation}
\end{figure*}

\paragraph*{Formalization} We also formalized this problem into a reinforcement learning problem. To find all the given items, the robot is given a list of items to collect, each one is represented by an image. The rest will be similar with the target-driven task. A first-person perspective image is given, an action is executed and then state is updated, reward is given.

To formalize that:
\begin{itemize}
    \item \textbf{Goal} The goal is also straight-forward, that the agent should find all the required object no matter the order.
    \item \textbf{Reward} The reward design would be similar with the target-driven navigation task though we would like to the finding of each object a separated but relatively smaller reward. We believe that such a design would be better for deep reinforcement training\cite{MJaderberg2016CoRR}. Also a minor negative time time punishment is necessary.
    \item \textbf{Action Space} Since the task type is similar with the target-driven problem. The same action space remains. \textbf{\{ MOVE FORWARD, MOVE BACKWARD, TURN LEFT, TURN RIGHT \}}
    \item \textbf{State Space} Different from the previous task, this task requires multiple images of objects to be perceived as targets. However, a set images with unknown-length is hard to be processed. Thus in this situation, we would limit the number of target inputs to 2 and the space of states would be the set of all possible package of three images
\end{itemize}

\paragraph*{Evaluation} Although the task is different from the previous task, the two major meters are the same. Hence, we use the same metric system in this task as is in the target-driven navigation task.

\subsection{Sweeper Route Planning}\label{subsec:applications/sweeper_route_planning}
\paragraph*{Motivation} Apart from target-driven navigation, there are also other valuable tasks in the field of indoor navigation. One widely-discussed but never well-addressed problem is the sweeper route planning problem. Contrary to the previous target-driven tasks, a sweeper robot should traverse over all the possible places in a given room without collision and less repetition. The current solutions often lead to stuck at some small space. Besides sweeper robot, it is general problem we will meet in many scenarios.

This problem is usually considered as coverage path planning (CPP) problem, which is usually dealt with using Boustrophedon method. However, the sweeper routing planning problem using visual semantic learning is actually very different from CPP problem. The most important factor is that the robot will not know that some movement is blocked until the collision really happens. It must see and understand to avoid collisions.

\begin{figure*}
\begin{center}
\fbox{\includegraphics[scale=1.0]{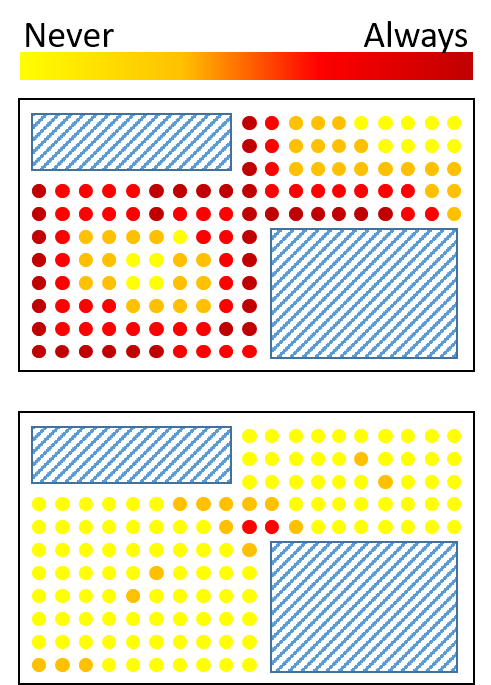}}
\end{center}
   \caption{Sweeper Route Planning (Upper: Random walking, Lower: DRL-based)}
\label{fig:applications/sweeper_route_planning/sweeping_robot}
\end{figure*}

\paragraph*{Formalization} Again, we formalized this problem into a reinforcement learning task. At any point, the robot is given an image, the first-person perception of the robot and an action should be given to the environment indicating which way the robot is going the next time slice. The environment should give out the reward then according to how much area has been covered by the sweeper. Typically, the robot should maintain a map of the room inside its system.

To formalize that:
\begin{itemize}
    \item \textbf{Goal} As a sweeper robot, surely the goal is to cover all the points without obstacles in least time. However, this goal is straight-forward but practically impossible to achieve as that many corners are hard to get and there is no need to sweep and to achieve a least time path it usually requires the god's perspective.
    \item \textbf{Reward} Reward design is simple in this problem. Since we do not pursuit a real all-cover path, we should reward every new position it gets and punish every fruitless movement. That is to say, we should give a large positive reward for a new position and a minor negative reward for an old position.
    \item \textbf{Action Space} In this task, a sweeper does not really care about which direction it is facing. The only thing it cares is that whether the place it is positioned and it will be positioned have been cleaned (went before). So we will not give the robot the ability to turn left/right. As the result, the action space shrinks to \textbf{\{ MOVE FORWARD, MOVE BACKWARD, MOVE LEFT, MOVE RIGHT \}}.
    \item \textbf{State Space} Different form the previous two tasks, sweeper path planning problem does not require a specific \emph{target} as an input so that the state space would be as simple as a set of all the images that the robot may see.
\end{itemize}

\paragraph*{Evaluation} The target for a sweeper is to sweep as much area as possible as fast as possible though these two aspect are usually contradict to each other. To deal with this, we defined a score similar to mAP score\cite{MEveringham2010IJCV} in object detection tasks to evaluate models. For the $n$ step in a trajectory, we define $m$ that in the first $n$ steps, $m$ points are covered. If the $n$ step is a new hit, a score of $m/n$ is given to the step, otherwise 0. An average over scores of all steps in a trajectory is the final score of the trajectory.

\begin{algorithm}
\caption{Mean Average Precision (mAP)}
\label{algo:mean_average_precision}
\begin{algorithmic}[H]
  \Function{mAP}{steps}
    \State{$score \gets 0$}
    \State{$n \gets 0$}
    \State{$m \gets 0$}
    
    \For{$step$ in $steps$}
      \State{$n \gets n + 1$}
      
      \If{$step$ is $NEW\_POSITION$}
        \State{$m \gets m + 1$}
        \State{$score \gets score + m/n$}
      \Else
        \State{$score \gets score + 0$}
        \EndIf
    \EndFor
    
    \State{$score \gets score / n$}
    
    \Return{$score$}
  \EndFunction
\end{algorithmic}
\end{algorithm}

\section{Auto-SLAM with Deep Reinforcement Learning}\label{sec:applications/autoslam_with_deep_reinforcement_learning}
\paragraph*{Motivation} Simultaneous localization and mapping (SLAM) problem has been paid much attention to since it is introduced decades ago. Many solutions have been proposed to increase the mapping accuracy. However, due to the lack of smart robots capable of navigating itself in alien environments, robots for SLAM tasks are still human controlled, which limited the application of SLAM technology significantly in dangerous places. Additionally, applying SLAM for some large-scale scenes (e.g. Museum) is effort consuming, which requires us to carry camera walked about all corners. So, we hope that robots can preform SLAM autonomously. With the help of NavigationNet, we proposed a Auto-SLAM task, to train a robot capable of navigating itself in an alien environment and complete the SLAM task without collision.

\paragraph*{Formalization} This task is also formalized as a reinforcement learning problem. On each step, the robot is fed with images perceived at its current location. The system could use the visual inputs and history information to improve the map. Reward should be calculated according to the difference between the constructed map and the ground truth. An action to lead the robot should be given on the basis of the reward and the current map, upon which the perception is updated.

To formally put it:
\begin{itemize}
    \item \textbf{Goal} Surely the goal is to build a 3D reconstruction efficiently and accurately. Though this would become a much more sophisticated problem than the other three due to the difficulties in evaluating the correctness of a new reconstruction.
    \item \textbf{Reward} Due to the difficulty of evaluation in SLAM problems, we would like to consider more about progress. By talking about progress, we are assuming total trust in legacy SLAM algorithms. We would like to give a positive reward for any new reconstructed area after each steps. In the meantime, a fruitless step is also punished with a minor negative reward.
    \item \textbf{Action Space} Due to the problem which will be talked about in \textbf{State Space} part, the function of turning-around is no longer required. In the meantime, we need to add the function of moving left/right to the robot. Hence the action space would become \textbf{\{ MOVE FORWARD, MOVE BACKWARD, MOVE LEFT, MOVE RIGHT \}}
    \item \textbf{State Space} Unlike the former three problems, SLAM is a problem far more than path planning. Most of it is about 3D reconstruction. Thus only a first-person single-eye perception is far from enough. In this setting, the environment should provide stereo vision from the four direction with depth information.
\end{itemize}

\paragraph*{Evaluation} SLAM tasks are divided into two part, mapping and positioning. Building maps is a way for positioning. Hence, we use the accuracy of positioning but not the of mapping as the benchmark. To evaluate a Auto-SLAM model, we ask the robot to run SLAM task in an alien enviroment with limited time. After that, 10 images are given as query to the positioning system. The score is calculated by average over the positioning success rate.

\section{Conclusion}\label{sec:conclusion}
We believe, to advance a research field, a suitable training environment and corresponding benchmarks will be the best catalyst. In this article, we proposed NavigationNet, a large-scale, open-source, low-cost, real-world  dataset for indoor navigation. By introducing this dataset, we hope to construct a platform where researchers can train and evaluate their` own robot controlling system without really constructing the robot. We extract robot controlling system out from robot itself. We hope to eliminate the long and high-cost preparation process before really training the system. Also, by constructing this dataset, we hope to make deep learning methods work on real robots. On NavigationNet, a robot will see as if it is in the physical world but the speed of action will become acceptable for massive trail-and-error.

We also proposed four possible tasks that can be conducted on our dataset. We are hoping that more tasks can be proposed to exploit its maximum potential.

Our future work includes increasing the number of scenes in the dataset. Also we will construct more attributes in the dataset (object segmentation for example).

%===============================================================================

% The maximum paper length is 8 pages excluding references and acknowledgements, and 10 pages including references and acknowledgements

\clearpage
% The acknowledgments are automatically included only in the final version of the paper.
%\acknowledgments{TODO}

%===============================================================================

\bibliography{main}  % .bib

\begin{thebibliography}{68}
\providecommand{\natexlab}[1]{#1}
\providecommand{\url}[1]{\texttt{#1}}
\expandafter\ifx\csname urlstyle\endcsname\relax
  \providecommand{\doi}[1]{doi: #1}\else
  \providecommand{\doi}{doi: \begingroup \urlstyle{rm}\Url}\fi

\bibitem[Liu et~al.(2007)Liu, Darabi, Banerjee, and Liu]{HLiu2007IEEESMC}
H.~Liu, H.~Darabi, P.~Banerjee, and J.~Liu.
\newblock Survey of wireless indoor positioning techniques and systems.
\newblock \emph{IEEE Transactions on Systems, Man, and Cybernetics, Part C
  (Applications and Reviews)}, 37:\penalty0 1067--1080, 2007.

\bibitem[Mnih et~al.(2013)Mnih, Kavukcuoglu, Silver, Graves, Antonoglou,
  Wierstra, and Riedmiller]{VMnih2013CoRR}
V.~Mnih, K.~Kavukcuoglu, D.~Silver, A.~Graves, I.~Antonoglou, D.~Wierstra, and
  M.~A. Riedmiller.
\newblock Playing atari with deep reinforcement learning.
\newblock \emph{CoRR}, abs/1312.5602, 2013.

\bibitem[Mnih et~al.(2016)Mnih, Badia, Mirza, Graves, Lillicrap, Harley,
  Silver, and Kavukcuoglu]{VMnih2016CoRR}
V.~Mnih, A.~P. Badia, M.~Mirza, A.~Graves, T.~P. Lillicrap, T.~Harley,
  D.~Silver, and K.~Kavukcuoglu.
\newblock Asynchronous methods for deep reinforcement learning.
\newblock \emph{CoRR}, abs/1602.01783, 2016.

\bibitem[Krizhevsky et~al.(2012)Krizhevsky, Sutskever, and
  Hinton]{KAlex2012NIPS}
A.~Krizhevsky, I.~Sutskever, and G.~E. Hinton.
\newblock Imagenet classification with deep convolutional neural networks.
\newblock In \emph{Proceedings of the 25th International Conference on Neural
  Information Processing Systems - Volume 1}, NIPS'12, pages 1097--1105, 2012.

\bibitem[He et~al.(2015)He, Zhang, Ren, and Sun]{KHe2015CoRR}
K.~He, X.~Zhang, S.~Ren, and J.~Sun.
\newblock Deep residual learning for image recognition.
\newblock \emph{CoRR}, abs/1512.03385, 2015.

\bibitem[Zhu et~al.(2016)Zhu, Mottaghi, Kolve, Lim, Gupta, Fei{-}Fei, and
  Farhadi]{YZhu2016CoRR}
Y.~Zhu, R.~Mottaghi, E.~Kolve, J.~J. Lim, A.~Gupta, L.~Fei{-}Fei, and
  A.~Farhadi.
\newblock Target-driven visual navigation in indoor scenes using deep
  reinforcement learning.
\newblock \emph{CoRR}, abs/1609.05143, 2016.

\bibitem[Song et~al.(2017)Song, Yu, Zeng, Chang, Savva, and
  Funkhouser]{SSong2016CVPR}
S.~Song, F.~Yu, A.~Zeng, A.~X. Chang, M.~Savva, and T.~Funkhouser.
\newblock Semantic scene completion from a single depth image.
\newblock \emph{IEEE Conference on Computer Vision and Pattern Recognition},
  2017.

\bibitem[Zhu et~al.(2017)Zhu, Gordon, Kolve, Fox, Fei{-}Fei, Gupta, Mottaghi,
  and Farhadi]{YZhu2017ICCV}
Y.~Zhu, D.~Gordon, E.~Kolve, D.~Fox, L.~Fei{-}Fei, A.~Gupta, R.~Mottaghi, and
  A.~Farhadi.
\newblock Visual semantic planning using deep successor representations.
\newblock \emph{{IEEE} International Conference on Computer Vision, {ICCV}
  2017}, pages 483--492, 2017.

\bibitem[Jang et~al.(2017)Jang, Vijaynarasimhan, Pastor, Ibarz, and
  Levine]{EJang2017CoRR}
E.~Jang, S.~Vijaynarasimhan, P.~Pastor, J.~Ibarz, and S.~Levine.
\newblock End-to-end learning of semantic grasping.
\newblock \emph{CoRR}, abs/1707.01932, 2017.

\bibitem[Gupta et~al.(2017)Gupta, Davidson, Levine, Sukthankar, and
  Malik]{SGupta2017CoRR}
S.~Gupta, J.~Davidson, S.~Levine, R.~Sukthankar, and J.~Malik.
\newblock Cognitive mapping and planning for visual navigation.
\newblock \emph{CoRR}, abs/1702.03920, 2017.

\bibitem[Huang et~al.(2016)Huang, Li, Zhang, He, Wu, Liu, Tang, and
  Zhuang]{SHuang2016IEEETIP}
S.~Huang, X.~Li, Z.~Zhang, Z.~He, F.~Wu, W.~Liu, J.~Tang, and Y.~Zhuang.
\newblock Deep learning driven visual path prediction from a single image.
\newblock \emph{IEEE Transactions on Image Processing}, 25\penalty0
  (12):\penalty0 5892--5904, 2016.

\bibitem[Finn et~al.(2015)Finn, Tan, Duan, Darrell, Levine, and
  Abbeel]{CFinn2015CoRR}
C.~Finn, X.~Y. Tan, Y.~Duan, T.~Darrell, S.~Levine, and P.~Abbeel.
\newblock Deep spatial autoencoders for visuomotor learning.
\newblock \emph{CoRR}, abs/1509.06113, 2015.

\bibitem[Canny(1988)]{JCanny1988CRMP}
J.~Canny.
\newblock \emph{The complexity of robot motion planning}.
\newblock MIT Press, 1988.

\bibitem[LaValle and Kuffner~Jr(2001)]{SLaValle2000RERT}
S.~M. LaValle and J.~J. Kuffner~Jr.
\newblock Rapidly-exploring random trees: Progress and prospects.
\newblock \emph{Algorithmic and Computational Robotics: New Directions}, 2001.

\bibitem[Sch\"{o}nberger and Frahm(2016)]{JSchonberger2016CVPR}
J.~L. Sch\"{o}nberger and J.-M. Frahm.
\newblock Structure-from-motion revisited.
\newblock \emph{IEEE Conference on Computer Vision and Pattern Recognition},
  2016.

\bibitem[Sch\"{o}nberger et~al.(2016)Sch\"{o}nberger, Zheng, Pollefeys, and
  Frahm]{JSchoenberger2016ECCV}
J.~L. Sch\"{o}nberger, E.~Zheng, M.~Pollefeys, and J.-M. Frahm.
\newblock Pixelwise view selection for unstructured multi-view stereo.
\newblock \emph{European Conference on Computer Vision (ECCV)}, 2016.

\bibitem[Pollefeys et~al.(2004)Pollefeys, Van~Gool, Vergauwen, Verbiest,
  Cornelis, Tops, and Koch]{MPollefeys2004IJCV}
M.~Pollefeys, L.~Van~Gool, M.~Vergauwen, F.~Verbiest, K.~Cornelis, J.~Tops, and
  R.~Koch.
\newblock Visual modeling with a hand-held camera.
\newblock \emph{International Journal of Computer Vision}, 59\penalty0
  (3):\penalty0 207--232, 2004.

\bibitem[Srivastava et~al.(2014)Srivastava, Fang, Riano, Chitnis, Russell, and
  Abbeel]{SSrivastava2014ICRA}
S.~Srivastava, E.~Fang, L.~Riano, R.~Chitnis, S.~Russell, and P.~Abbeel.
\newblock Combined task and motion planning through an extensible
  planner-independent interface layer.
\newblock \emph{Robotics and Automation (ICRA), 2014 IEEE International
  Conference on}, pages 639--646, 2014.

\bibitem[Zhu et~al.(2017)Zhu, Mottaghi, Kolve, Lim, Gupta, Fei-Fei, and
  Farhadi]{YZhu2017ICRA}
Y.~Zhu, R.~Mottaghi, E.~Kolve, J.~J. Lim, A.~Gupta, L.~Fei-Fei, and A.~Farhadi.
\newblock Target-driven visual navigation in indoor scenes using deep
  reinforcement learning.
\newblock \emph{Robotics and Automation (ICRA), 2017 IEEE International
  Conference on}, pages 3357--3364, 2017.

\bibitem[Levine et~al.(2016)Levine, Finn, Darrell, and Abbeel]{SLevine2016JMLR}
S.~Levine, C.~Finn, T.~Darrell, and P.~Abbeel.
\newblock End-to-end training of deep visuomotor policies.
\newblock \emph{Journal of Machine Learning Research}, 17\penalty0
  (39):\penalty0 1--40, 2016.

\bibitem[Kollar and Roy(2008)]{TKollar2008IJRR}
T.~Kollar and N.~Roy.
\newblock Trajectory optimization using reinforcement learning for map
  exploration.
\newblock \emph{The International Journal of Robotics Research}, 27\penalty0
  (2):\penalty0 175--196, 2008.

\bibitem[Fellbaum(1998)]{CFellbaum1998WN}
C.~Fellbaum.
\newblock \emph{WordNet: An Electronic Lexical Database}.
\newblock Bradford Books, 1998.

\bibitem[Russakovsky et~al.(2014)Russakovsky, Deng, Su, Krause, Satheesh, Ma,
  Huang, Karpathy, Khosla, Bernstein, Berg, and Li]{ORussakovsky2014CoRR}
O.~Russakovsky, J.~Deng, H.~Su, J.~Krause, S.~Satheesh, S.~Ma, Z.~Huang,
  A.~Karpathy, A.~Khosla, M.~S. Bernstein, A.~C. Berg, and F.~Li.
\newblock Imagenet large scale visual recognition challenge.
\newblock \emph{CoRR}, abs/1409.0575, 2014.

\bibitem[Bellemare et~al.(2012)Bellemare, Naddaf, Veness, and
  Bowling]{MBellemare2012CoRR}
M.~G. Bellemare, Y.~Naddaf, J.~Veness, and M.~Bowling.
\newblock The arcade learning environment: An evaluation platform for general
  agents.
\newblock \emph{CoRR}, abs/1207.4708, 2012.

\bibitem[Duan et~al.(2016)Duan, Chen, Houthooft, Schulman, and
  Abbeel]{YDuan2016CoRR}
Y.~Duan, X.~Chen, R.~Houthooft, J.~Schulman, and P.~Abbeel.
\newblock Benchmarking deep reinforcement learning for continuous control.
\newblock \emph{CoRR}, abs/1604.06778, 2016.

\bibitem[Brockman et~al.(2016)Brockman, Cheung, Pettersson, Schneider,
  Schulman, Tang, and Zaremba]{GBrockman2016arXiv}
G.~Brockman, V.~Cheung, L.~Pettersson, J.~Schneider, J.~Schulman, J.~Tang, and
  W.~Zaremba.
\newblock Openai gym, 2016.

\bibitem[Tanner and White(2009)]{BTanner2009JMLR}
B.~Tanner and A.~White.
\newblock Rl-glue: Language-independent software for reinforcement-learning
  experiments.
\newblock \emph{Journal of Machine Learning Research}, 10\penalty0
  (Sep):\penalty0 2133--2136, 2009.

\bibitem[Geramifard et~al.(2015)Geramifard, Dann, Klein, Dabney, and
  How]{AGeramifard2015JMLR}
A.~Geramifard, C.~Dann, R.~H. Klein, W.~Dabney, and J.~P. How.
\newblock Rlpy: a value-function-based reinforcement learning framework for
  education and research.
\newblock \emph{Journal of Machine Learning Research}, 16:\penalty0 1573--1578,
  2015.

\bibitem[Synnaeve et~al.(2016)Synnaeve, Nardelli, Auvolat, Chintala, Lacroix,
  Lin, Richoux, and Usunier]{GSynnaeve2016arXiv}
G.~Synnaeve, N.~Nardelli, A.~Auvolat, S.~Chintala, T.~Lacroix, Z.~Lin,
  F.~Richoux, and N.~Usunier.
\newblock Torchcraft: a library for machine learning research on real-time
  strategy games.
\newblock \emph{arXiv preprint arXiv:1611.00625}, 2016.

\bibitem[Beattie et~al.(2016)Beattie, Leibo, Teplyashin, Ward, Wainwright,
  K{\"u}ttler, Lefrancq, Green, Vald{\'e}s, Sadik, et~al.]{CBeattie2016arXiv}
C.~Beattie, J.~Z. Leibo, D.~Teplyashin, T.~Ward, M.~Wainwright, H.~K{\"u}ttler,
  A.~Lefrancq, S.~Green, V.~Vald{\'e}s, A.~Sadik, et~al.
\newblock Deepmind lab.
\newblock \emph{arXiv preprint arXiv:1612.03801}, 2016.

\bibitem[Kempka et~al.(2016)Kempka, Wydmuch, Runc, Toczek, and
  Ja{\'s}kowski]{MKempka2016CIG}
M.~Kempka, M.~Wydmuch, G.~Runc, J.~Toczek, and W.~Ja{\'s}kowski.
\newblock Vizdoom: A doom-based ai research platform for visual reinforcement
  learning.
\newblock \emph{Computational Intelligence and Games (CIG), 2016 IEEE
  Conference on}, pages 1--8, 2016.

\bibitem[Wymann et~al.(2014)Wymann, Espi{\'e}, Guionneau, Dimitrakakis, Coulom,
  and Sumner]{BWymann2014TORCS}
B.~Wymann, E.~Espi{\'e}, C.~Guionneau, C.~Dimitrakakis, R.~Coulom, and
  A.~Sumner.
\newblock Torcs, the open racing car simulator, 2014.

\bibitem[Todorov et~al.(2012)Todorov, Erez, and Tassa]{ETodorov2012IROS}
E.~Todorov, T.~Erez, and Y.~Tassa.
\newblock Mujoco: A physics engine for model-based control.
\newblock In \emph{Intelligent Robots and Systems (IROS), 2012 IEEE/RSJ
  International Conference on}, pages 5026--5033, 2012.

\bibitem[Chang et~al.(2017)Chang, Dai, Funkhouser, Halber, Niessner, Savva,
  Song, Zeng, and Zhang]{AChang20173DV}
A.~Chang, A.~Dai, T.~Funkhouser, M.~Halber, M.~Niessner, M.~Savva, S.~Song,
  A.~Zeng, and Y.~Zhang.
\newblock Matterport3d: Learning from rgb-d data in indoor environments.
\newblock \emph{International Conference on 3D Vision (3DV)}, 2017.

\bibitem[{Kolve} et~al.(2017){Kolve}, {Mottaghi}, {Gordon}, {Zhu}, {Gupta}, and
  {Farhadi}]{EKolve2017arXiv}
E.~{Kolve}, R.~{Mottaghi}, D.~{Gordon}, Y.~{Zhu}, A.~{Gupta}, and A.~{Farhadi}.
\newblock Ai2-thor: An interactive 3d environment for visual ai.
\newblock \emph{ArXiv e-prints}, 2017.

\bibitem[Sutton and Barto(1998)]{RSutton1998RL}
R.~S. Sutton and A.~G. Barto.
\newblock \emph{Reinforcement learning: An introduction}, volume~1.
\newblock MIT Press, 1998.

\bibitem[Mnih et~al.(2015)Mnih, Kavukcuoglu, Silver, Rusu, Veness, Bellemare,
  Graves, Riedmiller, Fidjeland, Ostrovski, et~al.]{VMnih2015Nature}
V.~Mnih, K.~Kavukcuoglu, D.~Silver, A.~A. Rusu, J.~Veness, M.~G. Bellemare,
  A.~Graves, M.~Riedmiller, A.~K. Fidjeland, G.~Ostrovski, et~al.
\newblock Human-level control through deep reinforcement learning.
\newblock \emph{Nature}, 518\penalty0 (7540):\penalty0 529--533, 2015.

\bibitem[Lillicrap et~al.(2015)Lillicrap, Hunt, Pritzel, Heess, Erez, Tassa,
  Silver, and Wierstra]{TLillicrap2015CoRR}
T.~P. Lillicrap, J.~J. Hunt, A.~Pritzel, N.~Heess, T.~Erez, Y.~Tassa,
  D.~Silver, and D.~Wierstra.
\newblock Continuous control with deep reinforcement learning.
\newblock \emph{CoRR}, abs/1509.02971, 2015.

\bibitem[Schulman et~al.(2015)Schulman, Moritz, Levine, Jordan, and
  Abbeel]{JSchulman2015CoRR}
J.~Schulman, P.~Moritz, S.~Levine, M.~I. Jordan, and P.~Abbeel.
\newblock High-dimensional continuous control using generalized advantage
  estimation.
\newblock \emph{CoRR}, abs/1506.02438, 2015.

\bibitem[Silver et~al.(2016)Silver, Huang, Maddison, Guez, Sifre, Van,
  Schrittwieser, Antonoglou, Panneershelvam, and Lanctot]{DSilver2016Nature}
D.~Silver, A.~Huang, C.~J. Maddison, A.~Guez, L.~Sifre, d.~D.~G. Van,
  J.~Schrittwieser, I.~Antonoglou, V.~Panneershelvam, and M.~Lanctot.
\newblock Mastering the game of go with deep neural networks and tree search.
\newblock \emph{Nature}, 529\penalty0 (7587):\penalty0 484, 2016.

\bibitem[Schulman et~al.(2015)Schulman, Levine, Moritz, Jordan, and
  Abbeel]{JSchulman2015CoRR2}
J.~Schulman, S.~Levine, P.~Moritz, M.~I. Jordan, and P.~Abbeel.
\newblock Trust region policy optimization.
\newblock \emph{CoRR}, abs/1502.05477, 2015.

\bibitem[Wu et~al.(2017)Wu, Mansimov, Liao, Grosse, and Ba]{YWu2017CoRR}
Y.~Wu, E.~Mansimov, S.~Liao, R.~B. Grosse, and J.~Ba.
\newblock Scalable trust-region method for deep reinforcement learning using
  kronecker-factored approximation.
\newblock \emph{CoRR}, abs/1708.05144, 2017.

\bibitem[Jaderberg et~al.(2016)Jaderberg, Mnih, Czarnecki, Schaul, Leibo,
  Silver, and Kavukcuoglu]{MJaderberg2016CoRR}
M.~Jaderberg, V.~Mnih, W.~M. Czarnecki, T.~Schaul, J.~Z. Leibo, D.~Silver, and
  K.~Kavukcuoglu.
\newblock Reinforcement learning with unsupervised auxiliary tasks.
\newblock \emph{CoRR}, abs/1611.05397, 2016.

\bibitem[Schulman et~al.(2017)Schulman, Wolski, Dhariwal, Radford, and
  Klimov]{JSchulman2017CoRR}
J.~Schulman, F.~Wolski, P.~Dhariwal, A.~Radford, and O.~Klimov.
\newblock Proximal policy optimization algorithms.
\newblock \emph{CoRR}, abs/1707.06347, 2017.

\bibitem[Kim et~al.(2004)Kim, Jordan, Sastry, and Ng]{HKim2004NIPS}
H.~J. Kim, M.~I. Jordan, S.~Sastry, and A.~Y. Ng.
\newblock Autonomous helicopter flight via reinforcement learning.
\newblock In \emph{Advances in neural information processing systems}, pages
  799--806, 2004.

\bibitem[Heess et~al.(2017)Heess, TB, Sriram, Lemmon, Merel, Wayne, Tassa,
  Erez, Wang, Eslami, Riedmiller, and Silver]{NHeess2017CoRR}
N.~Heess, D.~TB, S.~Sriram, J.~Lemmon, J.~Merel, G.~Wayne, Y.~Tassa, T.~Erez,
  Z.~Wang, S.~M.~A. Eslami, M.~A. Riedmiller, and D.~Silver.
\newblock Emergence of locomotion behaviours in rich environments.
\newblock \emph{CoRR}, abs/1707.02286, 2017.

\bibitem[Gu et~al.(2017)Gu, Holly, Lillicrap, and Levine]{SGu2017ICRA}
S.~Gu, E.~Holly, T.~Lillicrap, and S.~Levine.
\newblock Deep reinforcement learning for robotic manipulation with
  asynchronous off-policy updates.
\newblock In \emph{Robotics and Automation (ICRA), 2017 IEEE International
  Conference on}, pages 3389--3396, 2017.

\bibitem[Denil et~al.(2016)Denil, Agrawal, Kulkarni, Erez, Battaglia, and
  de~Freitas]{MDenil2016arXiv}
M.~Denil, P.~Agrawal, T.~D. Kulkarni, T.~Erez, P.~Battaglia, and N.~de~Freitas.
\newblock Learning to perform physics experiments via deep reinforcement
  learning.
\newblock \emph{arXiv preprint arXiv:1611.01843}, 2016.

\bibitem[Krizhevsky et~al.(2012)Krizhevsky, Sutskever, and
  Hinton]{AKrizhevsky2012NIPS}
A.~Krizhevsky, I.~Sutskever, and G.~E. Hinton.
\newblock Imagenet classification with deep convolutional neural networks.
\newblock In \emph{Advances in neural information processing systems}, pages
  1097--1105, 2012.

\bibitem[LeCun et~al.(1989)LeCun, Boser, Denker, Henderson, Howard, Hubbard,
  and Jackel]{YLeCun1989NC}
Y.~LeCun, B.~Boser, J.~S. Denker, D.~Henderson, R.~E. Howard, W.~Hubbard, and
  L.~D. Jackel.
\newblock Backpropagation applied to handwritten zip code recognition.
\newblock \emph{Neural computation}, 1\penalty0 (4):\penalty0 541--551, 1989.

\bibitem[Huang et~al.(2016)Huang, Liu, and Weinberger]{GHuang2016CoRR}
G.~Huang, Z.~Liu, and K.~Q. Weinberger.
\newblock Densely connected convolutional networks.
\newblock \emph{CoRR}, abs/1608.06993, 2016.

\bibitem[Chen et~al.(2017)Chen, Li, Xiao, Jin, Yan, and Feng]{YChen2017CoRR}
Y.~Chen, J.~Li, H.~Xiao, X.~Jin, S.~Yan, and J.~Feng.
\newblock Dual path networks.
\newblock \emph{CoRR}, abs/1707.01629, 2017.

\bibitem[Pang et~al.(2017)Pang, Sun, Ren, Yang, and Yan]{JPang2017arXiv}
J.~Pang, W.~Sun, J.~S. Ren, C.~Yang, and Q.~Yan.
\newblock Cascade residual learning: A two-stage convolutional neural network
  for stereo matching.
\newblock \emph{arXiv preprint arXiv:1708.09204}, 2017.

\bibitem[Zbontar and LeCun(2016)]{JZbontar2016JMLR}
J.~Zbontar and Y.~LeCun.
\newblock Stereo matching by training a convolutional neural network to compare
  image patches.
\newblock \emph{Journal of Machine Learning Research}, 17\penalty0
  (1-32):\penalty0 2, 2016.

\bibitem[Shaked and Wolf(2016)]{AShaked2016arXiv}
A.~Shaked and L.~Wolf.
\newblock Improved stereo matching with constant highway networks and
  reflective confidence learning.
\newblock \emph{arXiv preprint arXiv:1701.00165}, 2016.

\bibitem[Kendall et~al.(2017)Kendall, Martirosyan, Dasgupta, Henry, Kennedy,
  Bachrach, and Bry]{AKendall2017arXiv}
A.~Kendall, H.~Martirosyan, S.~Dasgupta, P.~Henry, R.~Kennedy, A.~Bachrach, and
  A.~Bry.
\newblock End-to-end learning of geometry and context for deep stereo
  regression.
\newblock \emph{arXiv preprint arXiv:1703.04309}, 2017.

\bibitem[Fuentes-Pacheco et~al.(2015)Fuentes-Pacheco, Ruiz-Ascencio, and
  Rend{\'o}n-Mancha]{JFuentesPacheco2015AIR}
J.~Fuentes-Pacheco, J.~Ruiz-Ascencio, and J.~M. Rend{\'o}n-Mancha.
\newblock Visual simultaneous localization and mapping: a survey.
\newblock \emph{Artificial Intelligence Review}, 43\penalty0 (1):\penalty0
  55--81, 2015.

\bibitem[Davison(2003)]{ADavison2003ICCV}
A.~J. Davison.
\newblock Real-time simultaneous localisation and mapping with a single camera.
\newblock In \emph{Proceedings Ninth IEEE International Conference on Computer
  Vision}, volume~2, pages 1403--1410, 2003.

\bibitem[Klein and Murray(2007)]{GKlein2007MAR}
G.~Klein and D.~Murray.
\newblock Parallel tracking and mapping for small ar workspaces.
\newblock In \emph{Mixed and Augmented Reality, 2007. ISMAR 2007. 6th IEEE and
  ACM International Symposium on}, pages 225--234, 2007.

\bibitem[Rublee et~al.(2011)Rublee, Rabaud, Konolige, and
  Bradski]{ERublee2011ICCV}
E.~Rublee, V.~Rabaud, K.~Konolige, and G.~Bradski.
\newblock Orb: An efficient alternative to sift or surf.
\newblock In \emph{Computer Vision (ICCV), 2011 IEEE international conference
  on}, pages 2564--2571, 2011.

\bibitem[Strasdat et~al.(2012)Strasdat, Montiel, and Davison]{HStrasdat2012IVC}
H.~Strasdat, J.~M. Montiel, and A.~J. Davison.
\newblock Visual slam: why filter?
\newblock \emph{Image and Vision Computing}, 30\penalty0 (2):\penalty0 65--77,
  2012.

\bibitem[Pire et~al.(2015)Pire, Fischer, Civera, De~Crist{\'o}foris, and
  Berlles]{TPire2015IROS}
T.~Pire, T.~Fischer, J.~Civera, P.~De~Crist{\'o}foris, and J.~J. Berlles.
\newblock Stereo parallel tracking and mapping for robot localization.
\newblock In \emph{Intelligent Robots and Systems (IROS), 2015 IEEE/RSJ
  International Conference on}, pages 1373--1378, 2015.

\bibitem[Mur-Artal and Tard\'os(2016)]{RMurArtal2016arXiv}
R.~Mur-Artal and J.~D. Tard\'os.
\newblock {ORB-SLAM2}: an open-source {SLAM} system for monocular, stereo and
  {RGB-D} cameras.
\newblock \emph{arXiv preprint arXiv:1610.06475}, 2016.

\bibitem[Whelan et~al.(2015)Whelan, Kaess, Johannsson, Fallon, Leonard, and
  McDonald]{TWhelan2015IJRR}
T.~Whelan, M.~Kaess, H.~Johannsson, M.~Fallon, J.~J. Leonard, and J.~McDonald.
\newblock Real-time large-scale dense rgb-d slam with volumetric fusion.
\newblock \emph{The International Journal of Robotics Research}, 34\penalty0
  (4-5):\penalty0 598--626, 2015.

\bibitem[Mur-Artal and Tard{\'o}s(2017)]{RMurArtal2017RAL}
R.~Mur-Artal and J.~D. Tard{\'o}s.
\newblock Visual-inertial monocular slam with map reuse.
\newblock \emph{IEEE Robotics and Automation Letters}, 2\penalty0 (2):\penalty0
  796--803, 2017.

\bibitem[Banzi(2008)]{MBanzi2008Arduino}
M.~Banzi.
\newblock \emph{Getting Started with Arduino}.
\newblock Make Books - Imprint of: O'Reilly Media, ill edition, 2008.

\bibitem[Halfacree and Upton(2012)]{GHalfacree2012RaspberryPi}
G.~Halfacree and E.~Upton.
\newblock \emph{Raspberry Pi User Guide}.
\newblock Wiley Publishing, 1st edition, 2012.

\bibitem[Everingham et~al.(2010)Everingham, Van~Gool, Williams, Winn, and
  Zisserman]{MEveringham2010IJCV}
M.~Everingham, L.~Van~Gool, C.~K.~I. Williams, J.~Winn, and A.~Zisserman.
\newblock The pascal visual object classes (voc) challenge.
\newblock \emph{International Journal of Computer Vision}, 88\penalty0
  (2):\penalty0 303--338, 2010.

\end{thebibliography}

\end{document}